\documentclass[twoside]{article}

\usepackage[preprint]{aistats2027}
\usepackage[round]{natbib}

\usepackage{tikz}
\usetikzlibrary{automata, positioning, arrows.meta, calc, decorations.pathreplacing}
\usepackage{amsmath,amssymb}
\usepackage{booktabs}
\newcommand{\tool}{SHIM}
\begin{document}



\twocolumn[

\aistatstitle{The Parser Already Knows: Lightweight Bias Correction in Constrained Decoding}

\aistatsauthor{ I\c{s}{\i}l \"{O}zg\"{u} \And Yaoxuan Wu \And Guy Van den Broeck \And Miryung Kim }

\aistatsaddress{ University of California, Los Angeles, Los Angeles, CA, USA \\ \texttt{\{isilozgu, thaddywu, guyvdb, miryung\}@cs.ucla.edu} } ]

\begin{abstract}
Grammar Constrained Decoding (GCD) forces Language Models (LMs) to produce syntactically valid outputs by masking out non-conforming tokens at each step. However, because masking only checks whether each token is valid so far, the resulting distribution over complete outputs diverges from the LM's own distribution conditioned on the grammar, biasing generation toward valid but suboptimal outputs. Online sampling can restore this distribution, but only through costly iterative resampling. Our key insight is that the parser and lexer states that GCD tools already maintain carry a strong signal about future grammatical validity. We introduce SHIM, a lightweight, offline-trained correction of the LM's next-token probabilities, conditioned on this syntactic and lexical state together with candidate next tokens. Since GCD tools already compute these states, SHIM leaves the LM itself untouched. Across bit-vector and text-to-SQL grammars, this correction substantially narrows the gap to the LM's grammar-conditioned distribution compared to masking and online sampling, while running at nearly masking's speed. Even a variant that sees only the next token can improve on both baselines, making SHIM usable with GCD tools that do not expose their parser state.
\end{abstract}

\section{Introduction}

\begin{figure}[t]
\centering
\definecolor{cbBlue}{RGB}{0,114,178}%    Okabe--Ito blue: sequences in L(alpha)
\definecolor{cbGreen}{HTML}{1B7837}%      dark green: SHIM correction
\definecolor{cbOrange}{RGB}{213,94,0}%   Okabe--Ito vermillion (masking distortion)
\begin{tikzpicture}[
    font=\scriptsize,
    >={Stealth[length=4pt]},
    one/.style={draw=white, line width=0.3pt, fill=cbBlue!85},
    zero/.style={draw=white, line width=0.3pt, fill=cbBlue!65!black},
    gone/.style={draw=white, line width=0.3pt, fill=black!13},
    rowlab/.style={anchor=east, align=right, inner sep=1pt},
    val/.style={anchor=center, inner sep=1pt},
    brace/.style={decorate, decoration={brace, amplitude=2.5pt, raise=1pt}, line width=0.4pt},
    hdr/.style={font=\scriptsize, inner sep=1pt},
    pl/.style={font=\bfseries\small, anchor=north west, inner sep=0pt},
    vc/.style={draw=black, line width=0.4pt, minimum width=1.12cm, minimum height=0.36cm,
               inner sep=0pt, outer sep=0pt},
]
% =====================================================================
% (a) Stacked bars: bar width = probability (common scale, width 1 = \W)
% =====================================================================
\def\W{4.0}   % bar length for total probability 1
\def\h{0.28}  % bar height
\def\VA{4.62} % x of value column "00000"
\def\VB{5.55} % x of value column "each 1****"
\def\X{1.55}  % x-offset of the bars (row labels sit left of it)
\begin{scope}[xshift=\X cm]
% ---------- LM p over all 32 strings ----------
\def\y{0}
\pgfmathsetmacro{\c}{\W/32}
\filldraw[zero] (0,\y) rectangle ++(\c,\h);
\foreach \i in {1,...,15}{\filldraw[gone] (\i*\c,\y) rectangle ++(\c,\h);}
\foreach \i in {16,...,31}{\filldraw[one] (\i*\c,\y) rectangle ++(\c,\h);}
\node[rowlab] at (-0.08,\y+\h/2) {LM $p(x)$};
\node[val] at (\VA,\y+\h/2) {$1/32$};
\node[val] at (\VB,\y+\h/2) {$1/32$};
\draw[brace] (\c,\y+\h) -- node[above=3pt, align=center, inner sep=0pt]
    {15 strings\\$\notin L(\alpha)$} (16*\c,\y+\h);
\draw[brace] (16*\c,\y+\h) -- node[above=3pt, align=center, inner sep=0pt]
    {16 strings \texttt{1****}\\$\in L(\alpha)$} (32*\c,\y+\h);
\draw[line width=0.4pt] (0.5*\c,\y+\h) -- (0.5*\c,\y+\h+0.66)
    node[above=-1pt, inner sep=1pt, font=\scriptsize\ttfamily] {00000};
\node[val, anchor=south, font=\scriptsize\ttfamily] at (\VA,\y+\h+0.06) {00000};
\node[val, anchor=south, align=center] at (\VB,\y+\h+0.06) {each\\\texttt{1****}};
\node[pl] at (-1.95,\y+\h+0.90) {(a)};

% ---------- target p(x|alpha) ----------
\def\y{-0.48}
\pgfmathsetmacro{\c}{\W/17}
\filldraw[zero] (0,\y) rectangle ++(\c,\h);
\foreach \i in {1,...,16}{\filldraw[one] (\i*\c,\y) rectangle ++(\c,\h);}
\node[rowlab] at (-0.08,\y+\h/2) {target $p(x\mid\alpha)$};
\node[val] at (\VA,\y+\h/2) {$1/17$};
\node[val] at (\VB,\y+\h/2) {$1/17$};

\draw[gray, dashed, line width=0.3pt] (-1.55,\y-0.12) -- (\VB+0.45,\y-0.12);

% ---------- token masking q_alpha ----------
\def\y{-1.00}
\pgfmathsetmacro{\c}{\W/32}
\filldraw[draw=white, line width=0.3pt, fill=cbOrange] (0,\y) rectangle ++(16*\c,\h);
\node[text=white, font=\scriptsize\ttfamily\bfseries] at (8*\c,\y+\h/2) {00000};
\foreach \i in {16,...,31}{\filldraw[one] (\i*\c,\y) rectangle ++(\c,\h);}
\node[rowlab] at (-0.08,\y+\h/2) {masking $q_\alpha(x)$};
\node[val, text=cbOrange] at (\VA,\y+\h/2) {$\boldsymbol{1/2}$};
\node[val, text=cbOrange] at (\VB,\y+\h/2) {$1/32$};

% ---------- SHIM q*_alpha ----------
\def\y{-1.48}
\pgfmathsetmacro{\c}{\W/17}
\filldraw[zero] (0,\y) rectangle ++(\c,\h);
\foreach \i in {1,...,16}{\filldraw[one] (\i*\c,\y) rectangle ++(\c,\h);}
\draw[cbGreen, line width=1.1pt] (0,\y) rectangle ++(\W,\h);
\node[rowlab, text=cbGreen] at (-0.08,\y+\h/2) {SHIM $q^*_\alpha(x)$};
\node[val, text=cbGreen] at (\VA,\y+\h/2) {${\approx}1/17$};
\node[val, text=cbGreen] at (\VB,\y+\h/2) {${\approx}1/17$};
\coordinate (abot) at (0,\y);
\end{scope}

% =====================================================================
% (b) Compact step-0 pipeline over x_1 in {0,1}
% =====================================================================
\path let \p1=(abot) in coordinate (B) at (0,\y1-0.32cm);
\node[pl] at ($(B)+(-0.40,0)$) {(b)};
\node[anchor=north west, inner sep=0pt] at ($(B)+(0.18,-0.02)$) {Step 0: choosing $x_1$ (prefix \texttt{""})};

% column x-positions
\def\xa{1.25}\def\xb{2.70}\def\xc{4.15}\def\xd{5.60}\def\xe{7.05}
\coordinate (H) at ($(B)+(0,-0.55)$);
\foreach \x/\t in {\xa/{$p(x_1)$},\xb/{mask},\xc/{$q_\alpha(x_1)$},\xd/{$\times\,\gamma(x_1)$},\xe/{$q^*_\alpha(x_1)$}}{
  \node[hdr] at ($(H)+(\x,0)$) {\t};
}
% row labels
\node[anchor=east, font=\scriptsize] at ($(H)+(0.62,-0.36)$) {$x_1{=}\texttt{0}$};
\node[anchor=east, font=\scriptsize] at ($(H)+(0.62,-0.72)$) {$x_1{=}\texttt{1}$};

\foreach \rr/\dy in {0/-0.36,1/-0.72}{
  \coordinate (Y\rr) at ($(H)+(0,\dy)$);
}
% cells
\node[vc] (P0) at ($(Y0)+(\xa,0)$) {$1/2$};
\node[vc] (P1) at ($(Y1)+(\xa,0)$) {$1/2$};
\node[vc] (M0) at ($(Y0)+(\xb,0)$) {valid};
\node[vc] (M1) at ($(Y1)+(\xb,0)$) {valid};
\node[vc, fill=cbOrange!12] (Q0) at ($(Y0)+(\xc,0)$) {$1/2$};
\node[vc, fill=cbOrange!12] (Q1) at ($(Y1)+(\xc,0)$) {$1/2$};
\node[vc, draw=cbGreen, fill=cbGreen!8] (G0) at ($(Y0)+(\xd,0)$) {$\approx 2/17$};
\node[vc, draw=cbGreen, fill=cbGreen!8] (G1) at ($(Y1)+(\xd,0)$) {$\approx 32/17$};
\node[vc, draw=cbGreen, fill=cbGreen!18, line width=0.9pt] (S0) at ($(Y0)+(\xe,0)$) {$\approx 1/17$};
\node[vc, draw=cbGreen, fill=cbGreen!18, line width=0.9pt] (S1) at ($(Y1)+(\xe,0)$) {$\approx 16/17$};
\foreach \i in {0,1}{
  \draw[->] (P\i) -- (M\i);
  \draw[->] (M\i) -- (Q\i);
  \draw[->] (Q\i) -- (G\i);
  \draw[->] (G\i) -- (S\i);
}

% SHIM feature box feeding gamma
\node[draw=cbGreen, rounded corners=2pt, fill=cbGreen!7, align=left, inner sep=2.5pt,
      anchor=north, font=\scriptsize] (F) at ($(G1.south)+(-1.3,-0.30)$)
  {\textbf{SHIM} (learned offline) reads parser state $S_0$,\\
   lexer state $L_0$, candidate next token $x_1$};
\draw[->, cbGreen, thick] (F.north -| G1.south) -- node[right, font=\scriptsize, text=cbGreen, inner sep=1pt] {$\gamma$} (G1.south);
\end{tikzpicture}
\caption{\textbf{Masking Skews the Distribution over Full Sequences, and SHIM's First-Step Correction Pulls It Back toward the Target.} $x$: full sequence; $x_1$: first token. \textbf{(a)}~Masking vs.\ the grammar-conditioned target distribution: masking gives \texttt{00000} probability $1/2$ instead of $1/17$. \textbf{(b)}~SHIM's learned correction at step~0.}
\label{fig:vector_progression_with_features}
\end{figure}

\paragraph{Problem Statement.} While Language Models (LMs) generate remarkably human-like text and code~\citep{JMLR:v24:22-1144, Li_2022, codex_chen2021evaluatinglargelanguagemodels, nl_to_code_austin2021programsynthesislargelanguage}, they consistently struggle to satisfy rigid syntactic and semantic constraints~\citep{lu2023boundingcapabilitieslargelanguage,li2025correctnessguaranteedcodegenerationconstrained}. In practice, prompt engineering and fine-tuning fall short of guaranteeing strict conformance to context-free grammars (CFGs)~\citep{lu2025learninggeneratestructuredoutput, tenckhoff2026llmstructbenchbenchmarkinglargelanguage}. This unreliability creates a bottleneck when integrating LMs into real-world software pipelines: even minor syntax violations yield malformed outputs that cause immediate downstream failures in parsers, input validators, structured data processing engines, and auto-formalization frameworks (e.g., SMT solvers or SQL engines).

\paragraph{Existing Approaches.} Prior work on constrained generation fundamentally trades generation quality for computational cost. The simplest approach, rejection sampling, repeatedly draws completions until a valid candidate emerges within a fixed budget~\citep{JMLR:v24:22-1144, chen2023teachinglargelanguagemodels, parys2026constrainedadaptiverejectionsampling}; however, this is computationally prohibitive and provides no formal compliance guarantee. Grammar Constrained Decoding (GCD) methods~\citep{ugare2025syncode,xgrammar_MLSYS2025_5c20ca4b,park2025flexibleefficientgrammarconstraineddecoding,outlines_willard2023efficientguidedgenerationlarge,domino_beurerkellner2024guidingllmsrightway}\footnote{Some prior work~\citep{lew2023sequentialmontecarlosteering,lipkin2025fastcontrolledgenerationlanguage,dang2026mitigatingbiaslocallyconstrained} calls this family \emph{Locally Constrained Decoding}; we use GCD throughout.} address this by masking non-conforming logits at each decoding step. While masking guarantees structural compliance with minimal overhead, it only evaluates whether a token is valid \emph{locally}, ignoring whether it forecloses high-probability completions down the line. Consequently, masking distorts the model's output distribution, often steering generation toward unnatural or degraded outputs~\citep{lew2023sequentialmontecarlosteering, asap_NEURIPS2024_2bdc2267}.

Online sampling techniques~\citep{asap_NEURIPS2024_2bdc2267,parys2026constrainedadaptiverejectionsampling,gonzalez2025constrainedsamplinglanguagemodels,lipkin2025fastcontrolledgenerationlanguage} restore this lost probability mass by dynamically reweighting candidates, but doing so requires expensive online resampling at every step, severely inflating inference latency. Alternatively, tractable probabilistic model augmentation~\citep{pmlr_v202_zhang23g,NEURIPS2024_d15c16cf,dang2026mitigatingbiaslocallyconstrained} avoids runtime sampling by pairing the LM with an offline surrogate model that is exactly computable. However, this requires learning a tractable approximation of the LM's full output distribution, which is costly and restricts the constraints it can express. Because each paradigm compromises on either latency, fidelity, or scalability, none of these options is ideal for practical deployment.

\paragraph{Our Solution.} To resolve this trilemma, we propose an offline, decoupled framework that achieves strict grammatical correctness without sacrificing inference speed. Rather than altering base model weights or performing expensive online resampling, we introduce SHIM: a lightweight, offline-trained probability controller. Like a physical shim aligning two mismatched surfaces, it aligns the masked distribution back to the LM's grammar-conditioned one, keeping generation natural and diverse rather than collapsing onto degenerate sequences. SHIM leaves the LM's internal computation untouched and only reweights its next-token probabilities; as Section~\ref{sec:methodology} shows, the ideal correction takes this form, a per-token factor on these probabilities, which SHIM learns to approximate. Our key insight is that the masking tool's own parser and lexer state~\citep{dragonbook_aho2006compilers} already carry much of the grammatical structure this correction needs. Instead of training a separate model to learn that structure from scratch, our controller reads it out directly, together with the candidate next token, without recomputing anything the masker has not already computed. In this way, \tool{} brings the constraint-aware lookahead of tractable-model methods to mainstream structured generation, fusing a small learned model with the classical parsing machinery GCD tools already run, and is fitted once per grammar so that every later request gets high-quality samples at near-masking speed. Masking alone ignores each token's effect on future validity, online sampling repeats its work for every request, and tractable-model methods must learn a surrogate of the full LM; \tool{} instead learns, once, only the correction that masking induces.

\paragraph{Intuition.} Masking tools decode with an incremental parser and lexer that track where the output stands in the grammar. When generating \texttt{(bvand s t)}, for instance, after the LM emits \texttt{(} and then \texttt{bv}, the parser has not yet moved past the opening parenthesis, but the lexer already knows that only \texttt{bvand}, \texttt{bvlshr}, and \texttt{bvor} remain possible. Such states carry a signal about which continuations stay valid after each candidate token, which masking alone ignores; \tool{} learns to read it out. Section~\ref{sec:methodology} walks through this example in detail.

\paragraph{Evaluation.} We implement \tool{} on top of Syncode~\citep{ugare2025syncode}, built on the Lark parsing toolkit~\citep{Lark_Parser_2018}, at three levels of feature availability and model capacity: a Multi-Layer Perceptron (MLP) and a logistic regression over the full feature set, and a logistic regression over the candidate next token alone. We measure performance via KL divergence against the target LM distribution and average inference latency, benchmarking SHIM against standard token masking via Syncode~\citep{ugare2025syncode} and the state-of-the-art online sampling framework ASAp~\citep{asap_NEURIPS2024_2bdc2267}. Averaged across grammars, our full-feature variants achieve roughly 4--5$\times$ lower KL divergence than masking and roughly 10$\times$ lower than online sampling, reaching up to two orders of magnitude on individual grammars, while correcting the distribution at a fraction of online sampling's cost: online sampling runs up to 20$\times$ slower than plain masking, whereas every SHIM variant stays within 1.25$\times$ of it.

\paragraph{Contributions.} (i) We show that the parser and lexer states a masker already computes carry a strong signal for correcting its distortion; (ii) we turn this into \tool{}, a plug-in controller that works with any masker exposing these states, needs no access to the LM's weights, and is fitted per model in minutes; and (iii) we evaluate it on bit-vector invariant grammars and on Spider text-to-SQL queries, a stress test on which online sampling's cost grows steeply while \tool{}'s stays near masking's.

\section{Characterizing the Probability Drift}
\label{sec:probability_drift}

Given an LM$p$, a prompt $x_{\leq s}$ of length $s$ and a grammar constraint $\alpha$, we want to sample from the conditional distribution 
\[
p(x_{>s} | x_{\leq s}, \alpha) = \prod_{i>s} p(x_{i} | x_{<i}, \alpha).
\]
Equivalently, this target is the LM's own distribution over output sequences, which no method (including ours) modifies, restricted to the grammar's support and renormalized:
\[
p(x_{>s} | x_{\leq s}, \alpha) = \frac{p(x_{>s} | x_{\leq s})\, \mathbb{I}(x_{>s} \in L(\alpha))}{\sum_{x'} p(x' | x_{\leq s})\, \mathbb{I}(x' \in L(\alpha))}.
\]
However, we do not have access to the conditional next-token distribution $p(x_{i} | x_{<i}, \alpha)$ and therefore this approach is not practical.
Instead, Grammar Constrained Decoding (GCD) tools give us access to a different distribution $q_\alpha$, which we can sample from directly. At the token level, $q_\alpha$ is defined as 
\[
q_\alpha(x_{i} | x_{< i}) = \frac{p(x_{i} | x_{< i}) \mathbb{I}(x_{< i}x_i \in \operatorname{Pref}(L(\alpha)))}{\sum_{t \in V} p(t | x_{< i})}
\]
 where $V$ stands for the grammatically valid next tokens for the prefix $x_{<i}$. $q_\alpha$ masks out any token that would make the prefix invalid and renormalizes the LM's probability over the tokens that remain valid, where a prefix $x_{< i}$ is valid if it can still be extended into a complete string in $L(\alpha)$. Over a full sequence, $q_\alpha$ factorizes the same way as the target:
\[
q_\alpha(x_{>s} | x_{\leq s}) = \prod_{i>s} q_\alpha(x_{i} | x_{<i}).
\]
Unfortunately, this approximation is not exact: $q_\alpha$ only masks based on whether the prefix conforms to the grammar \textit{so far}, i.e., \textit{past grammaticality}, so it does not incorporate the future grammaticality of the tokens it selects:
\[q_\alpha(x_{>s} | x_{\leq s}) \neq p(x_{>s} | x_{\leq s}, \alpha).\]

We demonstrate this with an example grammar $\mathcal{G}_{sk} = \texttt{00000} \mid \texttt{1}(\texttt{0}\mid\texttt{1})^4$, adapted from ASAp~\citep{asap_NEURIPS2024_2bdc2267} and illustrated step by step in Figure~\ref{fig:vector_progression_with_features}, given as a regular expression over tokens $\Sigma = \{0, 1\}$. Suppose the grammar accepts binary strings of length five, where a string can only continue with 0 if it started with 0. Suppose $p(x_i | x_{<i})$ is uniform. Simplifying the notation of $p(x_i | x_{<i})$ to $p(x)$ for sequence $x$, we have
$$p(00000 )=p(10000)=...=1/32$$
Then the target probability distribution $p(x \mid \alpha)$ for sequence $x$ should preserve the LM's distribution over the grammar-valid sequences, renormalized to sum to one, i.e., every grammar-satisfying string should have a probability of $1/17$:
$$p(00000| \alpha)=\frac{p(00000)}{\sum \limits_{x\in L(\alpha)} p(x)}
=\frac{1/32}{17/32}=1/17.$$

At each generation step $i$, GCD algorithms that use token-level masking would only sample the next token $x_{i}$ if $x_{\leq i}$ is a valid prefix of the grammar $\alpha$. However, this masked decoding algorithm would distort the final distribution. For example, assume that the LM would produce $x_{1}=0$ and $x_1=1$ with an even probability $1/2$. Suppose the masked LM chooses $x_{1}=0$. Then, because the only grammar-satisfying string that starts with 0 is 00000, the decoding algorithm would always append a 0 to the existing generated string, so the probability of generating $00000$ in the token-masked setting is 1/2. However, the actual probability should have been 1/17 to preserve the uniform distribution the LM initially had, because the LM models an even distribution over all 17 grammar-satisfying strings. Figure~\ref{fig:vector_progression_with_features}(a) shows this over all 32 sequences: 17 are grammatical and receive $1/17$ each under the target, while masking assigns \texttt{00000} alone $1/2$.

The root cause of the problem is that, in each decoding step, the algorithm should take into account the \textit{future}, or the \textit{probability of the next token yielding a valid sequence in the future}. Concretely, it should take into account the probability mass of valid future continuations, i.e., produce $x_{i}$ according to the probability $\sum \limits_{x_{>i}} p(x_{\geq i} \mid x_{<i})\, \mathbb{I}(x \in L(\alpha))$, which is intractable. Figure~\ref{fig:vector_progression_with_features}(b) shows how SHIM's correction factor $\gamma$ approximates this, pulling the masked, even $1/2$--$1/2$ split over $x_1$ toward the target $1/17$--$16/17$ split; after this first step no further correction is needed, since in this example the LM only emits five tokens from $\Sigma = \{0,1\}$, so every continuation after $x_1{=}1$ is grammar-valid and $x_1{=}0$ admits only \texttt{00000}.

\section{SHIM Methodology}
\label{sec:methodology}

At a high level, SHIM predicts a correction factor $\gamma$ for each candidate next token and applies it to the masked distribution $q_\alpha$ at every decoding step (Equation~\ref{eq:qstar}). We predict $\gamma$ from three features already exposed by the masking tool: the parser state, the lexer state, and the candidate next token. Below, we first derive what an ideal $\gamma$ would look like, simplify it into a tractable form, and then describe how we approximate it in practice with a lightweight model.

\subsection{Deriving the Ideal Correction}

\paragraph{Improving the Alignment.}
We aim to take $q_\alpha$, the distribution produced by an existing GCD tool, and improve it to become a better approximation $q^\star_\alpha$:
\[q^\star_\alpha(x_{>s} | x_{\leq s}) = \prod_{i>s} q^\star_\alpha(x_{i} | x_{<i}) \approx p(x_{>s} | x_{\leq s}, \alpha).\] 
Starting from an existing $q_\alpha$, we achieve this by applying a correction factor~$\gamma$ to the choice of next token: 
\begin{align} \label{eq:qstar}
    q^\star_\alpha(x_{i} | x_{<i}) \propto q_\alpha(x_{i} | x_{<i}) \cdot \gamma(x_{i}, x_{<i}),
\end{align}

\paragraph{An Impractical Solution.}
 If  $\gamma(x_{i}, x_{<i}) \approx \frac{p(x_{i} | x_{<i}, \alpha)}{q_\alpha(x_{i} | x_{<i})}$, then the following solution is correct but impractical.
\[q^\star_\alpha(x_{>s} | x_{\leq s}) \approx p(x_{>s} | x_{\leq s}, \alpha) \]

We can write $\gamma$ by using Bayes' rules and then simplifying the constant factors with respect to $x_i$,
{\small
\[
\gamma(x_{i}, x_{<i}) \approx \frac{ p(\alpha | x_{i}, x_{<i})\, p(x_{i} | x_{<i})}{q_\alpha(x_{i} | x_{<i}) \, p(\alpha | x_{<i})  }  \propto \frac{p(x_{i} | x_{<i}) \, p(\alpha | x_{i}, x_{<i})}{q_\alpha(x_{i} | x_{<i})}.
\]
}

\paragraph{Simplifying $\gamma$.}
Let $V = \left\{x_i \mid p(x_i | x_{< i}, \alpha) > 0\right\}$ be the set of tokens that can lead to a solution for $\alpha$. Then,
\begin{align*}
    q_\alpha(x_{i} | x_{< i}) = \begin{cases}
        0 \qquad \text{if~} x_i \not\in V \\
        \frac{p(x_{i} | x_{< i})}{\sum_{t \in V} p(t | x_{< i})} \qquad  \text{otherwise.}
    \end{cases}
\end{align*}
This means that the value of $\gamma$ is irrelevant for the first case, since the definition of $q^\star_\alpha$ multiplies it by zero.
It also allows us to simplify $\gamma$ for the second case. In that case, $q_\alpha(x_{i} | x_{< i}) \propto p(x_{i} | x_{<i})$ because $\sum_{t \in V} p(t | x_{< i})$ is a constant that does not depend on $x_i$ and the factor $\frac{p(x_{i} | x_{<i}) }{q_\alpha(x_{i} | x_{<i})}$ in $\gamma$ becomes a constant we can ignore. Thus, if  $\gamma(x_{i}, x_{<i}) \approx p(\alpha | x_{i}, x_{<i})$ we have that
\[q^\star_\alpha(x_{>s} | x_{\leq s}) \approx p(x_{>s} | x_{\leq s}, \alpha) .\]
In other words, the correction factor is the probability that the LM generates something that satisfies the constraint. Since the target $p(x_{>s} | x_{\leq s}, \alpha)$ is defined by the unmodified LM, this per-token factor on its output probabilities is all the correction requires; no change to the LM's internal states is needed.

\subsection{Learning the Correction}

\paragraph{Proposed Solution: Lightweight Grammar Alignment.}
Still, computing the quantity $p(\alpha | x_{i}, x_{<i})$ is intractable. A practical workaround is to train from data a regression model for $\gamma$ that looks at features of $(x_{i}, x_{<i})$ and predicts an estimate of $p(\alpha | x_{i}, x_{<i})$. Concretely, we approximate
\[
\gamma(x_{i}, x_{<i}) \approx \hat{\gamma}_\theta\big(S(x_{i}, x_{<i}),\, \ell(x_{i}, x_{<i}),\, x_{i}\big),
\]
where $S$ and $\ell$ are the parser and lexer states and $\hat{\gamma}_\theta$ is a small learned model. Both states are deterministic functions of the prefix: rather than omitting $x_{<i}$, they summarize the parts of it that bear on which continuations remain grammatical, discarding prefix details the grammar does not see; this is the source of the approximation. \tool{} is thus an amortized, state-conditioned estimate of the LM-weighted mass of valid continuations: prefixes that share these states may still be continued differently by the LM, and $\hat{\gamma}_\theta$ learns from samples the validity mass the LM typically assigns to continuations of each state.

\paragraph{Choosing the Features.} If $\alpha$ is a context-free grammar (CFG), we can use an interactive parser to expose its internal state $S(x_{i}, x_{<i})$ as a feature: this exposed state summarizes the prefix's grammaticality so far and correlates with which future tokens keep the sequence valid, making it a natural signal for the same future validity that $p(\alpha | x_{i}, x_{<i})$ measures. Moreover, since we apply $p(\alpha | x_{i}, x_{<i})$ as a probability correction to valid tokens, the next token itself is another intuitive feature: much like an LR(1) parser uses a single look-ahead token to resolve shift/reduce ambiguity about what to do next \citep{KNUTH1965607},  the candidate next token can itself trigger a parser state transition, making it an early signal of what comes next and whether the sequence stays grammatically valid. However, LM tokens and grammar lexemes rarely align one-to-one: a single LM token can complete only part of a lexeme, in which case the parser makes no progress and the parser state $S(x_{i}, x_{<i})$ stays unchanged, leaving it uninformative about what comes next. We therefore also expose the lexer state as a feature, since it tracks this finer-grained, within-lexeme progress and gives a signal about what the parser is likely to do once the lexeme completes, even when the parser state alone would not have moved. Parser state and lexer state are thus complementary: parser state captures structure across completed lexemes but stalls mid-lexeme, while lexer state captures within-lexeme progress but is blind to structure beyond it. Together with the next token, these three features give $\gamma$ a compact, informative signal of the sequence's grammatical trajectory. 

\begin{figure}[!tbp]
\centering
\begin{tikzpicture}[
    font=\sffamily\small,
    box/.style={draw=black!70, thick, rounded corners, fill=gray!5, inner sep=8pt}
]
\node[box] (grammar) {
    \begin{tabular}{l l l}
        \textbf{Start} & $::=$  & \texttt{s} $\mid$ \texttt{t} \\
                       & $\mid$ & \texttt{(} \textbf{Op} \textbf{Start} \textbf{Start} \texttt{)} $\mid\ \dots$ \\
        \textbf{Op}    & $::=$  & \texttt{bvand} $\mid$ \texttt{bvlshr} $\mid$ \texttt{bvor} $\mid\ \dots$
    \end{tabular}
};
\node[fill=gray!5, text=black, font=\bfseries\small, right=8pt] at (grammar.north west) {CFG: $\mathcal{G}_{\text{BV4}}$ excerpt};
\end{tikzpicture}
\caption{\textbf{Recursive Grammars Make It Hard to Anticipate Which Continuations Stay Valid.} Excerpt of the BV4 grammar used in our running example: \texttt{Start} recurses over binary operators applied to the inputs \texttt{s} and \texttt{t}.}
\label{fig:bv4_grammar}
\end{figure}

\begin{figure}[tp!]
\centering
\resizebox{\linewidth}{!}{%
\begin{tikzpicture}[
    shorten >=1pt,
    node distance=1.2cm and 1.3cm,
    auto,
    >={Stealth[length=3mm]},
    every state/.style={draw=black, thick, fill=white, minimum size=10mm, inner sep=1pt, font=\normalsize}
]

% --- Parser states while consuming the valid string "(bvand s t)" ---
\node[state] (S0) {$S_0$};
\node[state] (S1) [right=1.3cm of S0] {$S_1$};
\node[state] (S2) [right=2.1cm of S1] {$S_2$};
\node[state] (S3) [right=1.8cm of S2] {$S_3$};
\node[state] (S4) [right=1.8cm of S3] {$S_4$};
\node[state] (S5) [right=1.3cm of S4] {$S_5$};

\path[->] (S0) edge node {\texttt{(}} (S1);

% Op branch: bvand / bvlshr / bvor all collapse into the same parser state S2
\path[->] (S1) edge[bend left=25]  node {\texttt{bvand}} (S2);
\path[->] (S1) edge               node {\texttt{bvlshr}} (S2);
\path[->] (S1) edge[bend right=25] node[swap] {\texttt{bvor}} (S2);

% Start branch (1st operand): s / t collapse into the same parser state S3
\path[->] (S2) edge[bend left=15]  node {\texttt{s}} (S3);
\path[->] (S2) edge[bend right=15] node[swap] {\texttt{t}} (S3);

% Start branch (2nd operand): s / t collapse into the same parser state S4
\path[->] (S3) edge[bend left=15]  node {\texttt{s}} (S4);
\path[->] (S3) edge[bend right=15] node[swap] {\texttt{t}} (S4);

\path[->] (S4) edge node {\texttt{)}} (S5);

\node[below=0.3cm of S0, font=\small, text=gray, align=center] {Start};
\node[below=0.3cm of S2, font=\small, text=gray, align=center] {Op consumed};
\node[below=0.3cm of S5, font=\small, text=gray, align=center] {Accept};

\end{tikzpicture}%
}
\caption{\textbf{The Parser State May Lag behind Generation When LM Tokens Split Lexemes.} Parser state transitions while decoding \texttt{(bvand s t)} under the grammar of Figure~\ref{fig:bv4_grammar}; the parser advances only once a lexeme completes, and all three operators lead to the same state $S_2$.}\label{fig:binary_grammar_dfa_fixed_final}
\end{figure}

\paragraph{A Worked Example.} Structured generation tools like Syncode maintain exactly this kind of incremental parser and lexer state~\citep{incremental_parsing_10.1145/293677.293678} as they decode; we illustrate how these states work, and how our controller uses them, with a concrete example. Consider parsing the expression \texttt{(bvand s t)} under the BV4 grammar excerpt in Figure~\ref{fig:bv4_grammar}. The opening \texttt{(} is consumed as its own token, advancing the parser to $S_1$ in Figure~\ref{fig:binary_grammar_dfa_fixed_final}; suppose the LM's tokenizer then produces \texttt{b} and \texttt{v} as its next two tokens. The LM itself generates at the token level, but the lexer state tracks progress at the character level: its \emph{buffer} accumulates the characters of the current, in-progress lexeme only, emptying each time a lexeme completes -- unlike the prefix $x_{<i}$, which accumulates every token the LM has generated. Since \texttt{(} already completed as its own lexeme, the buffer excludes it and accumulates only \texttt{b} and \texttt{v}, regardless of how these characters were chunked into tokens. Right after these two tokens, the buffer is \texttt{"bv"}, landing the lexer state on node $v$ in Figure~\ref{fig:lexical_search_space}. Formally, the lexer state is the set of lexemes that remain consistent with this buffer, together with how far it has advanced into each candidate lexeme's pattern. Here, \texttt{bvand}, \texttt{bvlshr}, and \texttt{bvor} are all still possible. The next token then narrows this set to the intended lexeme.

\begin{figure}[tp!]
\centering
\resizebox{\linewidth}{!}{%
\begin{tikzpicture}[
    font=\sffamily, >=Stealth,
    node distance=1.5cm and 2cm,
    circ/.style={circle, draw=black!80, thick, fill=white, minimum size=0.8cm, align=center},
    valid/.style={circle, draw=green!60!black, thick, fill=green!10, minimum size=0.8cm},
    invalid/.style={circle, draw=red!60!black, thick, fill=red!10, minimum size=0.8cm},
    current/.style={circle, draw=teal!80!black, thick, fill=teal!10, minimum size=0.8cm, double},
    arrow/.style={->, thick},
    reject/.style={->, thick, draw=red!60!black, dashed}
]

% Nodes
\node[circ] (root) at (0, 0) {$\epsilon$};
\node[circ] (b) at (2, 0) {\texttt{b}};
\node[current] (v) at (4, 0) {\texttt{v}};

% Adjusted Y-coordinates for tighter spacing
\node[valid] (a) at (6, 1.5) {\texttt{a}};
\node[valid] (l) at (6, 0.5) {\texttt{l}};
\node[valid] (o) at (6, -0.5) {\texttt{o}};
\node[invalid] (x) at (6, -1.5) {\texttt{x}};

% Adjusted Y-coordinates for corresponding labels
\node[font=\footnotesize, text=gray] (and) at (8.3, 1.5) { \texttt{nd} (\texttt{BVAND})};
\node[font=\footnotesize, text=gray] (shr) at (8.5, 0.5) { \texttt{shr} (\texttt{BVLSHR})};
\node[font=\footnotesize, text=gray] (or)  at (8.1, -0.5) { \texttt{r} (\texttt{BVOR})};

% Edges
\draw[arrow] (root) -- (b);
\draw[arrow] (b) -- (v);

\draw[arrow, draw=green!60!black] (v) -- (a);
\draw[arrow, draw=green!60!black] (v) -- (l);
\draw[arrow, draw=green!60!black] (v) -- (o);
\draw[reject] (v) -- (x) node[midway, below left, font=\scriptsize, text=red!60!black] {Masked / Rejected};

\draw[arrow, dotted, text=gray] (a) -- (and);
\draw[arrow, dotted, text=gray] (l) -- (shr);
\draw[arrow, dotted, text=gray] (o) -- (or);

% Annotations
\node[above=1.2cm of v, xshift=-1.1cm, text=teal!80!black, font=\bfseries\small, align=center] {Lexer State at Step $t$ \\ (Buffer: \texttt{"bv"})};

\end{tikzpicture}%
}
\caption{\textbf{Mid-Lexeme, the Lexer State and the Candidate Token Carry the Signal before the Parser Catches Up.} The lexer state is a node in this terminal trie: buffer \texttt{"bv"} cannot distinguish \texttt{bvand}, \texttt{bvlshr}, and \texttt{bvor}, but the candidate next token (e.g., \texttt{"a"}, \texttt{"l"}, \texttt{"o"}) does; invalid tokens (e.g., \texttt{"x"}) are masked.
}\label{fig:lexical_search_space}
\end{figure}

At this point in decoding, the parser itself is still at state $S_1$ in Figure~\ref{fig:binary_grammar_dfa_fixed_final}: it only advances once a complete operator token is consumed, so it gives no indication that the possibilities have already narrowed to \texttt{bvand}, \texttt{bvlshr}, and \texttt{bvor}. Only the lexer state captures this within-lexeme progress; the parser catches up once the operator token completes and it advances to $S_2$.

\paragraph{Training Data.} We draw 1000 unconstrained LM completions per grammar, independently of the evaluation samples, replay each through the masker's incremental parser, and record one row per generated token: the parser state, the lexer state, and the token the LM produced. All rows of a completion share its label, whether the finished output parses, so the controller directly estimates $p(\alpha \mid x_{i}, x_{<i})$. Since the GCD tool exposes these states as discrete identifiers, we one-hot encode the recent parser states, the lexer state, the partially matched lexeme, and the candidate token, and concatenate them.

\paragraph{Models.} We use two capacities: a logistic regression, cheap to train and to evaluate at each step, and, to capture non-linear feature interactions, a small Multi-Layer Perceptron (MLP) with two hidden layers of 64 and 32 ReLU units, chosen among small architectures by cross-validation.

At inference, we add $\log\hat{\gamma}_\theta$ to the masked logits of each allowed token before renormalizing, so the correction requires no iterative sampling. \tool{} thus functions as a refinement layer on top of widely used masking tools such as Syncode~\citep{ugare2025syncode}, XGrammar~\citep{xgrammar_MLSYS2025_5c20ca4b}, and Outlines~\citep{outlines_willard2023efficientguidedgenerationlarge}; XGrammar and Outlines, for instance, are structured-output backends in the SGLang serving engine~\citep{zheng2024sglang}, and XGrammar also in vLLM~\citep{kwon2023vllm}. Because it only reweights tokens the masker already allows, it preserves the masker's grammatical guarantee. It also needs only the masker's own parser and lexer, so a different LM or tokenizer changes the mask and \tool{}'s inputs together; retrained on that LM's samples, \tool{} inherits the masker's applicability by construction, while making a masker work for a new grammar or tokenizer is an orthogonal problem.
\begin{figure*}[!t]
    \centering
    \includegraphics[width=1\textwidth]{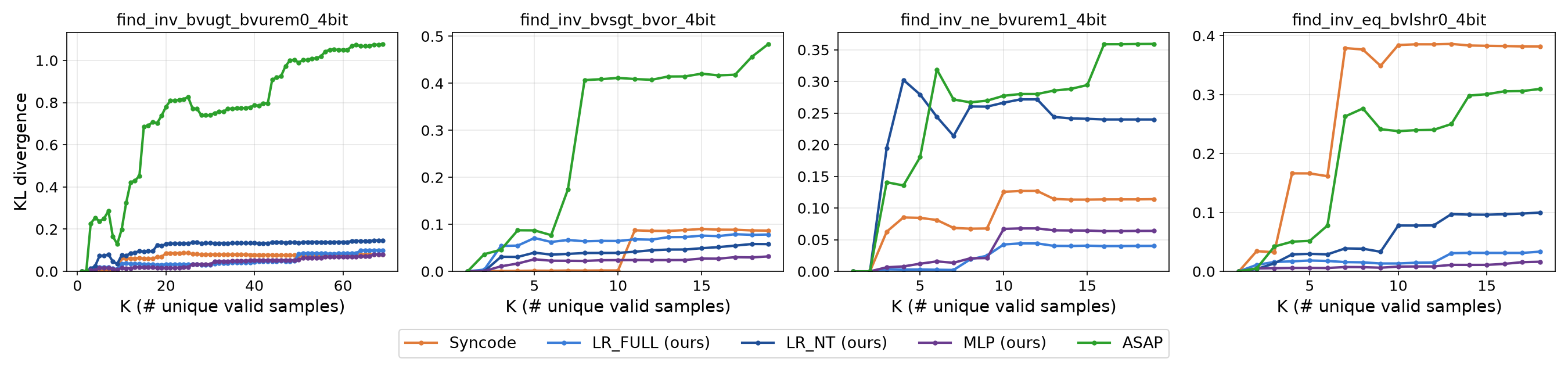}
\caption{\textbf{SHIM's Advantage Holds as More of the Target Is Revealed.} KL divergence on four representative grammars as a function of $k$, the number of unique valid LM samples used to estimate the target; our full-feature variants stay low as $k$ grows, while ASAp's divergence rises.}    \label{fig:reverse_kl_grid}
\end{figure*}

\section{Experiments}

\paragraph{Baselines.} We compare SHIM against Syncode~\citep{ugare2025syncode}, a state-of-the-art masking-based GCD tool, and ASAp~\citep{asap_NEURIPS2024_2bdc2267}, an online sampling method that iteratively reweights its distribution toward the target. We use ASAp as the representative of online sampling: CARS and MCMC belong to the same sample-and-correct family, and in a cost-only pilot on our BV4 and SQL tasks they share its overhead: MCMC is 2.4--20$\times$ slower than \tool{} per valid sample, and CARS takes 3.4--4.2$\times$ longer to return its first valid sample while rejecting 44--62\% of its attempts. Tractable augmentation methods such as Ctrl-G instead require the constraint as a finite automaton paired with a distilled HMM, so they cannot directly handle general context-free grammars such as SQL's. Because ASAp's resampling cost grows with each sample, running it to convergence was computationally infeasible at our evaluation scale, so we cap it at 100 samples per grammar, a budget at which it is already by far the slowest method we evaluate. SHIM's cost, by contrast, is paid once per grammar and offline: its 1,000 training sequences come from the unconstrained LM, each costing one ordinary LM sample (Table~\ref{tab:ana-rev-results}) and generated independently, so they can be batched, and fitting the controller takes at most five minutes, so across repeated use it quickly becomes cheaper than online sampling, whose overhead recurs for every sample.

\paragraph{Metrics.} We measure probabilistic integrity via the KL divergence $D_{KL}(p(\cdot\mid\alpha)\,\|\,q)$ from the grammar-conditioned target to a method's distribution $q$, and computational efficiency via average end-to-end inference time per sample, which reflects both a method's per-step cost and the length of the outputs it produces. We compute it on the valid sequences the LM itself produces: we draw 1000 samples from the LM, keep the grammar-valid ones as the support $S$, and renormalize both distributions over $S$, using each sequence's exact probability under the LM and under each method. This measures how faithfully a method preserves the LM's relative preferences among its own valid outputs, and is applied identically to every method. Figure~\ref{fig:reverse_kl_grid} shows that \tool{}'s estimates remain stable as $S$ grows.

\paragraph{Experimental Setup.} We evaluate on the Syntax Guided Synthesis (SyGuS) invariant-generation problems~\citep{alur2019syguscomp2018resultsanalysis}, following prior work~\citep{asap_NEURIPS2024_2bdc2267}, using \texttt{Mistral-7B-Instruct-v0.2}~\citep{jiang2023mistral} as the base LM. We focus on INV-BV, SyGuS's bit-vector invertibility-condition split and the same benchmark ASAp was evaluated on; its grammar variants span a range of difficulty over a small, closed vocabulary, which keeps the target distribution tractable to estimate from samples. We additionally evaluate on text-to-SQL using Spider~\citep{yu2019spiderlargescalehumanlabeleddataset}, whose schema-dependent identifiers form a much larger, open vocabulary than INV-BV's, probing whether the gains carry over to this setting; it also serves as a stress test for online sampling, whose cost grows steeply on such grammars. Table~\ref{tab:ana-rev-results} gives the complete picture across all methods and BV4 grammars, and we draw our conclusions from trends across these grammars rather than from any single row; the remainder of this section breaks this down into three parts, mirroring the feature levels introduced previously: (1) a full-feature MLP correction using parser state, lexer state, and the candidate next token, (2) a lightweight logistic regression on the same features, compared against Syncode and our online sampling baseline ASAp, and (3) a next-token-only correction. All methods share the same prompt and sample with temperature 1.0, top-p 1.0 and no top-k truncation, so that sampling follows the LM's own distribution; generation is capped at 80 new tokens for BV4 and 200 for SQL. We will release our code and generated training data publicly upon publication under an open-source license.

\paragraph{(1) Full-Feature Correction with an MLP.} We first ask whether a corrector with access to the full feature set (next token, parser state, and lexer state) can recover more of the LM's original distribution than masking alone. Table~\ref{tab:ana-rev-results} reports the results for Syncode, our MLP variant, and ASAp across all 14 grammars: our MLP variant reduces KL divergence over Syncode on 11 of 14 grammars and over ASAp on all 14, running roughly $4.6\times$ faster than ASAp on average (3.64s vs.\ 16.62s). It remains slower than Syncode, though (3.64s vs.\ 2.97s) -- the overhead we address next with a lighter-weight model.

\paragraph{(2) A Lightweight Logistic Regression Matches the MLP at a Fraction of the Cost.} We next replace the MLP with a lightweight logistic regression (\texttt{LR\_FULL}) trained on the same full feature set, and compare both against Syncode and ASAp. Table~\ref{tab:ana-rev-results} shows that \texttt{LR\_FULL} matches or beats Syncode on 12 of 14 grammars, which is almost the same coverage as the MLP variant. ASAp, despite reweighting its sampling distribution online, never achieves the lowest KL divergence in this comparison.

Figure~\ref{fig:reverse_kl_grid} shows whether these gaps hold as more of the target is revealed, plotting KL against $k$, the number of unique valid LM samples. A small-$k$ estimate can hide gaps, so a rising curve reflects a more complete view of the target rather than a method degrading. \texttt{MLP} and \texttt{LR\_FULL} stay low and flat, \texttt{LR\_NT} stabilizes, and ASAp climbs steadily, ending worst on three of the four grammars: its competitiveness at small $k$ reflects under-sampling.

This gap widens once we account for inference cost. Table~\ref{tab:ana-rev-results}'s bottom row shows \texttt{LR\_FULL} running in 2.52s on average -- faster than Syncode's own 2.97s -- and roughly $7\times$ faster than ASAp's 16.62s. Because we train \texttt{LR\_FULL} only once, offline, it requires no per-sample online training, unlike ASAp, which must update its internal representation after every generated sequence. Computing $\gamma$ adds a small per-step cost, but by correcting for future grammaticality, \texttt{LR\_FULL} likely avoids the forced, longer completions that plain masking's short-sighted choices can produce; the resulting savings from needing fewer steps outweigh that added cost.

\begin{table}[!t]
\centering
\small
\caption{KL Divergence and Latency on BV4 Grammars. KL divergence ($\downarrow$) to the grammar-conditioned target per grammar, and average latency per sequence (seconds). Our logistic-regression variants even outpace masked decoding, while online sampling takes over 5$\times$ as long.}\label{tab:ana-rev-results}
\setlength{\tabcolsep}{1.55pt}
\begin{tabular}{l|cc|ccc}
\toprule
 & \multicolumn{2}{c|}{Baselines} & \multicolumn{3}{c}{Ours (SHIM)} \\
\cmidrule(lr){2-3}\cmidrule(lr){4-6}
Grammar & Syncode & ASAp & MLP & \shortstack{LR\_\\FULL} & \shortstack{LR\_\\NT} \\
\midrule
 eq\_bvlshr0 & 0.382 & 0.310 & \textbf{0.016} & 0.033 & 0.100 \\
 bvsge\_bvashr1 & 0.522 & 0.583 & 0.139 & \textbf{0.082} & 0.200 \\
 ne\_bvudiv1 & 0.248 & 0.197 & 0.0023 & \textbf{0.0022} & 0.012 \\
 ne\_bvudiv0 & 0.079 & 0.239 & 0.017 & \textbf{0.003} & 0.107 \\
 ne\_bvurem1 & 0.114 & 0.359 & 0.064 & \textbf{0.040} & 0.240 \\
 ne\_bvurem0 & 0.311 & 0.168 & \textbf{0.008} & 0.024 & 0.097 \\
 bvsge\_bvnot & 0.006 & 0.148 & 0.008 & \textbf{0.001} & 0.177 \\
 bvugt\_bvneg & 0.277 & 0.386 & \textbf{0.033} & 0.117 & 0.282 \\
 ne\_bvneg & 0.032 & 0.235 & 0.003 & \textbf{0.001} & 0.015 \\
 bvsge\_bvneg & 0.046 & 0.167 & \textbf{0.007} & 0.014 & 0.196 \\
 bvugt\_bvurem1 & \textbf{0.016} & 0.308 & 0.075 & 0.031 & 0.094 \\
 bvugt\_bvurem0 & \textbf{0.080} & 1.078 & 0.082 & 0.100 & 0.146 \\
 bvsgt\_bvor & 0.087 & 0.484 & \textbf{0.032} & 0.078 & 0.058 \\
eq\_bvand & 0.104 & 0.258 & 0.003 & \textbf{0.002} & 0.080 \\
\midrule
\textbf{Avg. time} (LM: 3.40) & 2.97 & 16.62 & 3.64 & 2.52 & \textbf{2.45} \\
\bottomrule
\end{tabular}
\end{table}

\paragraph{(3) The Next Token Alone Can Help, Even with No Parser or Lexer Access.} Finally, \texttt{LR\_NT} restricts the corrector to \emph{only} the candidate next token. It is a weaker corrector than the full-feature variants: it beats Syncode on exactly half the grammars, and when it loses the gap can be large (e.g.\ bvsge\_bvnot), but it needs no parser or lexer state and runs in 2.45s on average, slightly faster than Syncode and roughly $7\times$ faster than ASAp. Where it helps, it is because BV4's terminals are short: individual LM tokens often complete a lexeme and trigger a parser transition, so the next token acts as an implicit lookahead (Section~\ref{sec:methodology}). Comparing \texttt{LR\_NT} with \texttt{LR\_FULL} also serves as an ablation of the state features: collapsing the representation to the next token alone raises average KL by roughly 3$\times$, indicating that the parser and lexer states carry much of the future-validity signal. Where the next token alone is already informative, as on bvsgt\_bvor, \texttt{LR\_NT} can still edge out \texttt{LR\_FULL}.

\paragraph{(4) A Second Domain: Text-to-SQL.} We use a sample of four pairs from Spider as our benchmark. Table~\ref{tab:spider_kl_results} reports KL divergence for each row. As on BV4, our lightweight correctors match or improve on Syncode's masked distribution on most rows (store\_product, entertainment\_awards, cre\_theme\_park), though Syncode remains best on election.

The \texttt{LR\_NT} column shows that the next-token benefit from Part~(3) is conditional. \texttt{LR\_NT} wins on store\_product, whose \texttt{LIKE "\%Scanner\%"} pattern makes the next token nearly the whole signal once the LM starts spelling it out, but is worst on entertainment\_awards, whose near-duplicate column names (e.g.\ \texttt{Num\_of\_Audience}, \texttt{NumOfAudience}) diverge only many tokens in, exactly the bookkeeping the parser and lexer state provide. Election, the only \texttt{JOIN} row, is its second-worst, since which table a column belongs to is a structural decision the next token cannot track.

Generation time diverges sharply as well. On store\_product, Syncode and our correctors take 1.7--4.0s per valid sample while ASAp takes $18.87$s, and ASAp's cost grows more steeply on the other rows, reaching $93$--$108$s on election and cre\_theme\_park; the latter required simplifying ASAp's vocabulary, which changes the target, so we omit that row's KL. On the full, unsimplified SQL grammar, ASAp needed about 84 minutes for a single sample and its internal state grew by roughly 7--8 MiB per generated token, whereas \tool{} samples in 3--4 seconds with a fixed controller of under 2 MiB: ASAp tracks every live parse alternative separately, so its cost can grow combinatorially with repetitive vocabularies, while \tool{}'s per-step cost is fixed.

\begin{table}[t]
\centering
\small
\caption{KL Divergence and Latency on Spider Text-to-SQL. Rows report KL divergence per query; the bottom row shows mean inference time (seconds). $^{\S}$Includes cre\_theme\_park, whose grammar we simplified so ASAp's sampling terminates; this can only lower ASAp's reported time.}
\setlength{\tabcolsep}{2pt}
\begin{tabular}{l|cc|ccc}
\toprule
 & \multicolumn{2}{c|}{Baselines} & \multicolumn{3}{c}{Ours (SHIM)} \\
\cmidrule(lr){2-3}\cmidrule(lr){4-6}
Grammar (row) & Syncode & ASAp & MLP & \shortstack{LR\_\\FULL} & \shortstack{LR\_\\NT} \\
\midrule
store\_product & 0.418 & 0.717 & 0.310 & 0.327 & \textbf{0.128} \\
entertainment\_awards & 0.082 & 0.128 & \textbf{0.026} & 0.030 & 1.986 \\
election & \textbf{0.251} & 0.300 & 0.431 & 0.375 & 0.746 \\
cre\_theme\_park & 0.060 & --- & \textbf{0.041} & 0.049 & 0.295 \\
\midrule
\textbf{Avg. time} (LM: 5.16) & \textbf{3.86} & 80.44$^{\S}$ & 4.54 & 4.78 & 3.95 \\
\bottomrule
\end{tabular}
\label{tab:spider_kl_results}
\end{table}

\section{Related Work}

\paragraph{Grammar Constrained Decoding.} Masking-based GCD tools differ mainly in how they make masking efficient: Syncode~\citep{ugare2025syncode}, XGrammar~\citep{xgrammar_MLSYS2025_5c20ca4b}, and GREATGRAMMA~\citep{park2025flexibleefficientgrammarconstraineddecoding} precompute token masks or token--lexeme maps, Outlines~\citep{outlines_willard2023efficientguidedgenerationlarge} compiles the grammar into an automaton, and DOMINO~\citep{domino_beurerkellner2024guidingllmsrightway} aligns LM tokens with grammar subterminals. All check conformance only for tokens generated \emph{so far}, the source of the distortion we study; \tool{} keeps their masks and reuses their parser and lexer state to correct it.

\paragraph{Online Sampling Methods.} ASAp~\citep{asap_NEURIPS2024_2bdc2267} and CARS~\citep{parys2026constrainedadaptiverejectionsampling} record grammatical dead ends and reweight later samples away from them; MCMC~\citep{gonzalez2025constrainedsamplinglanguagemodels} and AWRS~\citep{lipkin2025fastcontrolledgenerationlanguage} refine or adaptively reject proposals; and SMC Steering~\citep{lew2023sequentialmontecarlosteering} frames constrained generation as sequential Monte Carlo inference. These methods target the grammar-conditioned distribution directly, but must update their state as they sample, so their cost recurs with every sample. \tool{} instead fixes its correction offline and evaluates only a small model per step.

\paragraph{Tractable Probabilistic Model Augmentation.} GeLaTo~\citep{pmlr_v202_zhang23g}, Ctrl-G~\citep{NEURIPS2024_d15c16cf}, and P-GCD~\citep{dang2026mitigatingbiaslocallyconstrained} pair the LM with a tractable model, typically an HMM distilled from LM samples, whose constraint-satisfaction likelihood reweights each next-token distribution. GeLaTo, Ctrl-G, and P-GCD differ in how they represent constraints, using CNF keyword clauses, DFAs, and tensorized automata combined with SMC, respectively. \tool{} carries this idea of constraint-aware lookahead into mainstream structured generation: instead of distilling a surrogate for the LM's full output distribution, it fuses a small learned model with the parser and lexer state that GCD tools already maintain. It targets the common setting where a grammar is reused across many requests, so a one-time offline fit pays off in every later sample. FUDGE~\citep{yang2021fudge} and NADO~\citep{meng2022nado} learn similar predictors for general controlled generation; \tool{} specializes them to grammars while keeping masking's validity guarantee.

\section{Conclusion}

%dont critise older work, focus on what we did. %kl divergence numbers
We introduced \tool{}, which corrects the distortion that grammar-constrained masking introduces into an LM's output distribution. Rather than resampling online or distilling a surrogate of the LM, \tool{} learns a small offline controller over signals that masking tools already compute but, to our knowledge, have not previously used for correction: the parser and lexer states, together with the candidate next token. Applied as a per-token reweighting of the masked distribution, it runs at nearly the speed of plain masking. Evaluated on both bit-vector invariant grammars and text-to-SQL queries, its full-feature variants bring the output 4--5$\times$ closer to the target in KL divergence than masking does on BV4, and about 10$\times$ closer than online sampling, while improving on masking for most of our Spider queries; even a controller that sees only the next token narrows the gap on average. Overall, offline, state-conditioned correction delivers distributional fidelity at close to masking's speed.

\bibliography{references}

\end{document}